%% file: mlsp_template_camera_ready.tex
\documentclass{article}
\usepackage{amsmath,graphicx,mlspconf}
\usepackage{xcolor}

\usepackage{microtype}
\usepackage{graphicx}
\usepackage{subfigure}
\usepackage{booktabs} 

\usepackage{amsmath}
\usepackage{amssymb}
\usepackage{mathtools}
\usepackage{amsthm}
\usepackage{multirow}

\usepackage{hyperref}
\usepackage[capitalize,noabbrev]{cleveref}
\usepackage{xcolor}

\theoremstyle{plain}

\theoremstyle{definition}

\theoremstyle{remark}

\usepackage{algorithm}
\usepackage{algorithmic}
\usepackage{forloop}

\input{math_commads}

\copyrightnotice{979-8-3195-0884-3/26/\$31.00 {\copyright}2026 IEEE}

\toappear{2026 IEEE International Workshop on Machine Learning for Signal Processing, Sep.\ 28-- Oct.\ 1, 2026, Atlanta, USA}

\title{Pheno-GS: Phenoscape-scale Geodesic Sinkhorn}

\name{%
Alistair Wilkinson$^{\star \dagger}$%
    \qquad Christopher J. Tape$^{\star}$%
    \qquad Smita Krishnaswamy$^{\dagger}$
}
\address{%
    $^{\star}$ Department of Oncology, University College London, London, UK. \\%
    $^{\dagger}$ Department of Computer Science, Yale University, New Haven, CT, USA. %
}

\begin{document}

\maketitle

\begin{abstract}
High-throughput single-cell data is now collected across large patient cohorts. Understanding patient-level heterogeneity from cellular-level data motivates phenoscaping: embedding each single-cell distribution as a "datapoint," with distances given by optimal transport (OT). Computing geometry-aware OT at this scale, between all pairs of patient datasets, remains an open challenge, since existing methods either rely on Euclidean ground metrics that distort manifold structure or fail under sparse, unevenly sampled, or large-scale data. We present \textbf{Pheno-GS} (Phenoscape-scale Geodesic Sinkhorn), which computes accurate, scalable geodesic transport distances under noisy, unbalanced, large-scale settings via three components: ($1$) graph connectivity regularization for well-defined geodesics on sparse/disconnected manifolds; ($2$) an unbalanced OT formulation via KL marginal penalties; and ($3$) a batched matrix algorithm computing all pairwise distances in one heat diffusion (over $200 \times$ faster than Geodesic Sinkhorn for $500$ distributions). We validate Pheno-GS on synthetic benchmarks and a CyTOF perturbation dataset.

\end{abstract}
\section{Introduction}
Single-cell technologies, such as CyTOF, scRNA-seq, and scATAC-seq, are now routinely used to phenotype hundreds-to-thousands of perturbation conditions. Each condition comprises a cell-by-gene (or cell-by-protein) matrix, i.e., a high-dimensional distribution. A recent approach to analyze such data is the phenoscape: an OT-based map of samples that supports downstream clustering or classification \cite{chen2020uncovering, tong2021diffusion, huguet2023geodesic}. Existing methods, however, ($1$) ignore the underlying manifold structure of the data, ($2$) scale poorly to large cohorts, and ($3$) can yield accurate distances without realistic transport plans. This is a limitation that matters biologically, since the transition from a healthy to a sick patient reflects a disease-progression path along which cells die or proliferate, requiring unbalanced OT. We address all three challenges with Phenoscape-scale Geodesic Sinkhorn (\textbf{Pheno-GS})\footnote{\url{https://github.com/KrishnaswamyLab/PhenoGS}}.

Pheno-GS builds on scalable entropy-regularized OT, solved via Sinkhorn iterations, while respecting data geometry by modeling the data as a graph with explicit bridges for connectivity. It extends this to an unbalanced formulation via KL marginal penalties, and further addresses scalability with a batched matrix algorithm that computes all pairwise distances in a single heat diffusion rather than $n^2$ sequential runs. The graph connectivity module underlies both the unbalanced and batched variants, giving a common foundation for robust geodesic computation across all operating modes. We validate Pheno-GS on synthetic manifold benchmarks and a single-cell patient-derived organoid (PDO) CyTOF phenoscape dataset.

\section{Related Work}\vspace{-2mm}
GS builds on the link between entropy-regularized OT and diffusion on geometric domains where the heat operator efficiently approximates geodesic distances on meshes ~\cite{crane_geodesics_2013, solomon2015convolutional}, though sparse Cholesky solvers still scale as $\mathcal{O}(n^3)$ in the worst case. DiffusionEMD ~\cite{tong2021diffusion} embeds distributions via graph diffusion operators but does not compute OT distances directly, trading accuracy for speed, as our benchmarks confirm. Unbalanced OT (UOT) relaxes strict marginal constraints via KL penalties. Pham et al. ~\cite{pham2020unbalanced} show entropic UOT's Sinkhorn algorithm finds an $\varepsilon$-approximate solution in  $\mathcal{O}(n^2/\varepsilon)$ time, better than the $\mathcal{O}(n^2/\varepsilon^2)$ rate for classical Sinkhorn.

\section{Preliminaries}
OT compares distributions $\boldsymbol{\mu}, \boldsymbol{\nu}$ on $(\mathcal{X}, d)$ by minimizing transport cost. The 2-Wasserstein distance,
\[
W_2(\boldsymbol{\mu},\boldsymbol{\nu}) := \left(\inf_{\pi \in \Pi(\boldsymbol{\mu},\boldsymbol{\nu})} \int_{\mathcal{X}^2} d(x,y)^2\,\mathrm{d}\pi(x,y)\right)^{1/2},
\]

costs $\mathcal{O}(n^3\log n)$. Adding a KL penalty yields the Sinkhorn distance~\cite{cuturi2013sinkhorn}, solvable in $\mathcal{O}(n^2/\varepsilon^2)$ via iterative rescaling of $K_\lambda(x,y)=e^{-d(x,y)^2/\lambda}$~\cite{dvurechensky_computational_2018}.
For manifold-structured data, Euclidean distances distort intrinsic geometry. Following~\cite{huguet2023geodesic}, $K_\lambda$ is replaced by the heat kernel $\mathcal{H}_t$, which encodes geodesic distances via Varadhan's formula~\cite{varadhan_behavior_1967},
\[
d(x,y)^2 = \lim_{t\to 0^+} -4t\ln\mathcal{H}_t(x,y),
\]

giving heat-geodesic distance ${d_{\mathcal{H}}}^2(x,y) := -4t\ln\mathcal{H}_t(x,y)$ and kernel $K_\lambda = \mathcal{H}_{\lambda/4}$.
Huguet et al. show that, for all $t > 0$, the Harnack bound gives $h_t(x,y) \approx V(x,\sqrt{t})^{-1} e^{\left({-d(x,y)^2/4t}\right)}$, which inverts to $d(x,y)^2 \approx -4t \log h_t(x,y) - 4t \log V(x,\sqrt{t})$, providing a usable finite-t heat-geodesic dissimilarity.
On a weighted graph $\mathcal{G}=(V,E)$ with Laplacian $\mL=\mD-\mA$, the heat operator $\mH_t = e^{-t\mL}$ is efficiently approximated using degree-$K$ Chebyshev polynomials~\cite{shuman_chebyshev_2011, marcotte2022fast}. Given that the maximum eigenvalue satisfies $\lambda_{max}(L) = 2$, it follows that,
\[
\mH_t \approx p_K(\mL,t) = \tfrac{c_0}{2}\mI + \sum_{k=1}^{K} c_k T_k(\mL-\mI),
\]

at $\mathcal{O}(Kmn)$ for an $m$-nearest-neighbor graph. The resulting Sinkhorn updates for distributions $\vmu, \vnu$ on $\mathcal{G}$, derived in~\cite{huguet2023geodesic}, are
\begin{equation}\label{eq:geo_sinkhorn_updates}
    \vv \leftarrow \vmu \oslash p_K(\mL, t)(\va \odot \vw),
    \qquad
    \vw \leftarrow \vnu \oslash p_K(\mL, t)(\va \odot \vv),
\end{equation}

avoiding the dense $n\times n$ kernel entirely. However GS assumes $\mathcal{G}$ is fully connected. Disconnected components make $\mH_t$ block-diagonal, preventing diffusion across components and yielding undefined geodesic distances.
\noindent \section{Pheno-GS}\vspace{-2mm}

Pheno-GS comprises an unbalanced module (UGS) and a batched matrix module (BMS), both built on a shared graph connectivity module (GC) that can ensure well-defined geodesic distances under sparse or disconnected graph structure. Together they constitute a unified framework for robust, phenoscape-scale manifold OT.


\subsection{Graph Connectivity Regularization Module (GC)}

While graph-based geodesic Sinkhorn provides optimal transport paths penalized by a manifold- or graph-path-based cost for a connected graph, its behavior is undefined on a disconnected graph as graph diffusions are not defined between truly disconnected components. To circumvent this, most graph-based OT methods, including~\cite{huguet2023geodesic}, connect the graph via a fully-connected weighted adjacency matrix built from a Gaussian or similar kernel. However, such kernels have exponential drop-off in affinity as distance increases, so for a Gaussian kernel any pair of points more than one standard deviation apart has near-zero affinity. Fully-connected graphs built this way thus fail to distinguish disconnected-but-nearby manifolds from disconnected-and-far manifolds. We instead propose adding a linear bridge between disconnected manifolds, effectively using linear Euclidean distance in the disconnected case.

Given $X \subset \mathbb{R}^d$, we select a subset $S$ of $M$ points by uniform subsampling or $k$-means centroids, then generate $N_a$ Gaussian auxiliary points around each $s_i \in S$ with covariance $\sigma^2 I$, forming $\mathcal{N} = \{n_{ij}\}$. We construct an affinity matrix between auxiliary points,
\[
[\mK_{SN}]_{ij} = \exp\!\left(-\frac{\|s_i - n_j\|^2}{2\sigma^2}\right), \qquad \mK_{NN} = \mK_{SN}^\top K_{SN},
\]
and compute the row-stochastic transition matrix $\mP_{NN} = \mD^{-1} \mK_{NN}$. Running $t$ steps of diffusion and projecting back onto $\mN$ yields manifold-aligned auxiliary points: $\mN_{\text{adj}} = \mP_{NN}^t \mN$. The augmented dataset $\mX' = \mX \cup \mN_{\text{adj}}$ then constructs a connected graph (Algorithm~\ref{alg:graph_connectivity}).

\begin{figure}[!t]
    \centering
    \includegraphics[width=\linewidth]{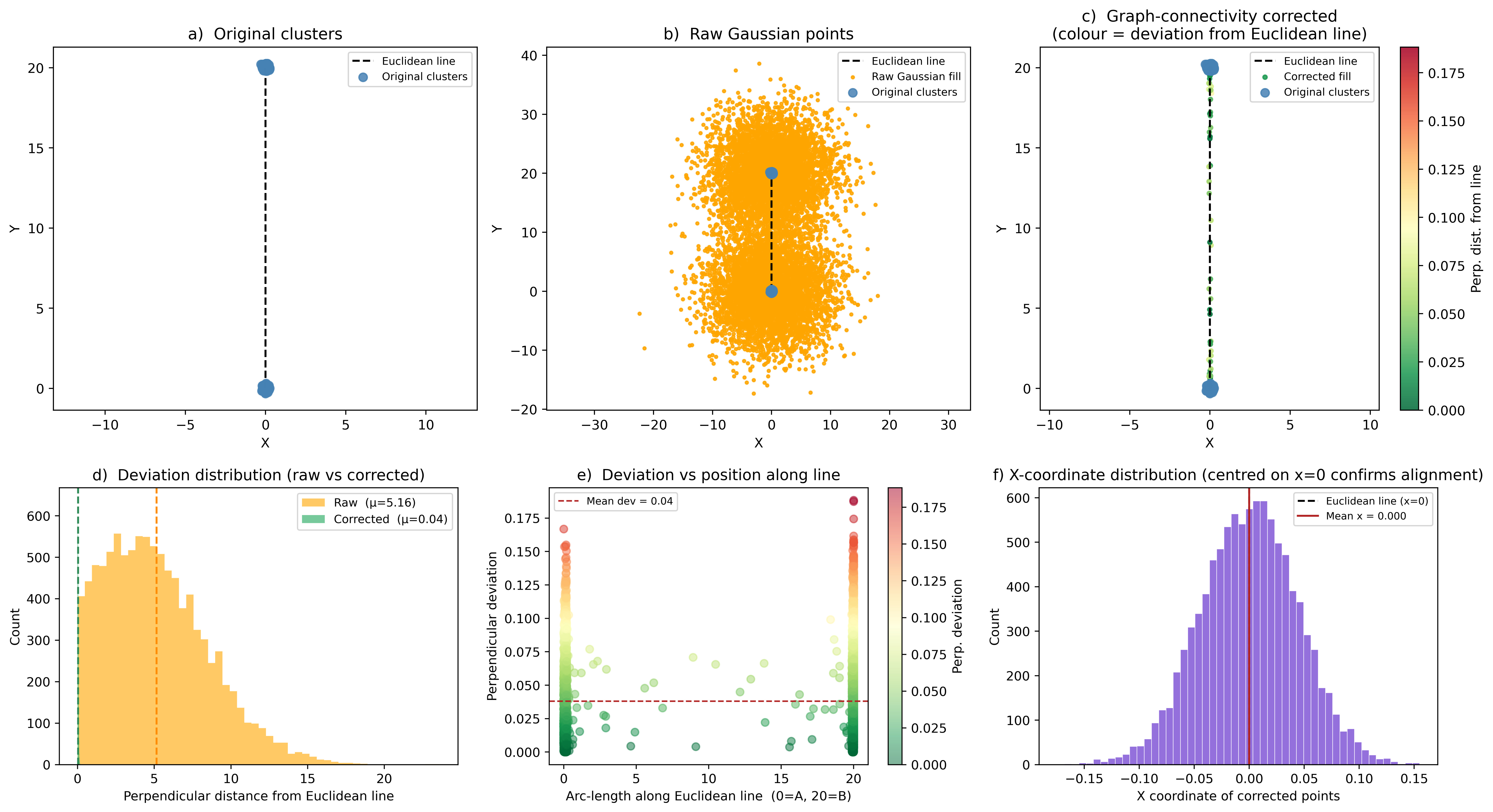}
    \caption{\footnotesize Validation of graph connectivity regularization on a synthetic two-cluster dataset. \textbf{(a)} Original clusters with Euclidean geodesic (dashed). \textbf{(b)} Raw Gaussian auxiliary points scattered broadly in ambient space. \textbf{(c)} After diffusion, auxiliary points collapse onto the geodesic. \textbf{(d)} Deviation distributions before and after correction. \textbf{(e)} Perpendicular deviation vs.\ arc-length along the geodesic. \textbf{(f)} $x$-coordinate distribution of corrected points. }
    \label{fig:line_connection}
\end{figure}

A simpler fix, such as increasing $k$ or the kernel bandwidth $\sigma$, can yield a connected graph but risks linking points on opposing manifold surfaces, introducing spurious shortcuts that distort geodesic distances. 
By contrast, the diffusion $\mN_{adj} = \mP_{NN}^t \mN$ replaces each auxiliary point by a random-walk-weighted mean of its neighbors, $[\mN_{adj}]_j = \sum_i (\mP^t)_{ji} \boldsymbol{n}_i$, pulling points toward regions of high local density. For two disconnected components $O_A$, $O_B$ with a sparsely sampled gap between them, auxiliary points seeded off the inter-cluster axis are pulled back toward the nearer high-density surface, while points seeded near the axis are pulled toward the sparse but non-empty gap region, so the surviving auxiliaries concentrate along the low-dimensional path of highest density connecting the components. Fig.~\ref{fig:line_connection} confirms this empirically: after diffusion the perpendicular deviation from the Euclidean geodesic $O_A$ $O_B$ collapses ($5.16 \rightarrow 0.04$).

GC is one-time preprocessing ($t=1$) adding $P = M N_a$ auxiliary nodes ($M$ anchors, $N_a$ auxiliaries per anchor). Forming $\mK_{NN}$ costs $\mathcal{O}(M \cdot P^2)$; degree scaling plus one diffusion step costs $\mathcal{O}(P^2 d)$ in dimension $d$, giving $\mathcal{O}(P^2(M+d))$ overall.

\begin{algorithm}[!t]
    \caption{Graph Connectivity Regularization}
    \label{alg:graph_connectivity}
\begin{algorithmic}
    \footnotesize
    \STATE {\bfseries Input:} Dataset $\mX$, number of anchor points $M$, auxiliary points per anchor $N_a$, diffusion steps $t$
    \STATE {\bfseries Output:} Augmented dataset $\mX' = \mX \cup \mN_{\text{adj}}$
    \STATE Sample $M$ anchor points $\mS \subset \mX$
    \STATE Generate Gaussian auxiliary points $\mN$ centered at each $s_i \in \mS$
    \STATE Construct $\mK_{SN}$ (anchors to auxiliaries); set $\mK_{NN} = \mK_{SN}^\top \mK_{SN}$
    \STATE Compute $\mP_{NN} = \mathrm{diag}(\mK_{NN}\mathbf{1})^{-1}\mK_{NN}$
    \STATE $\mN_{\text{adj}} \leftarrow \mP_{NN}^t\,\mN$
    \STATE {\bfseries Return:} $\mX' = \mX \cup \mN_{\text{adj}}$
\end{algorithmic}
\end{algorithm}
\begin{figure}[!t]
    \centering
    \includegraphics[width=\linewidth]{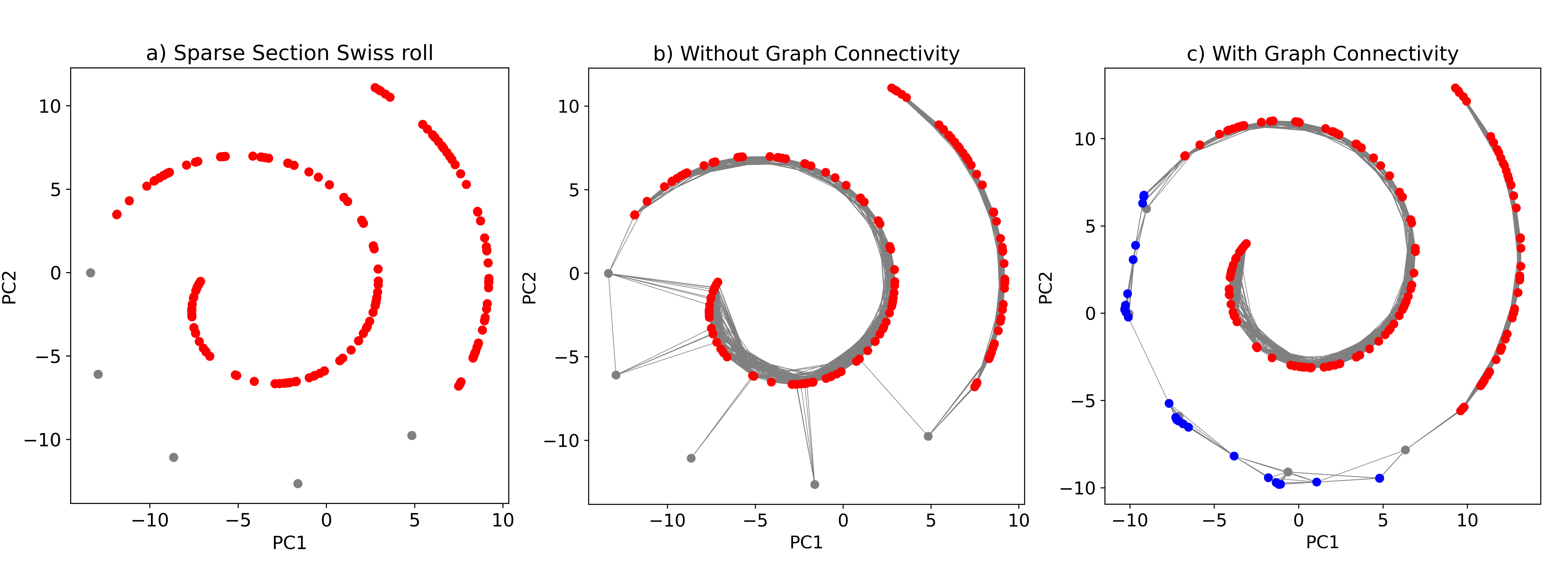}
    \caption{\footnotesize Graph connectivity regularization on synthetic data. \textbf{(a--c)} Sparse Swiss roll ($100$ points, $10$ graph-components): original distorted manifold graph, Gaussian auxiliary points generated over the sparse section ($\sigma=0.75$, $N_a=25$ per anchor). The manifold-aligned auxiliaries yield a single connected component.}
    \label{fig:swiss_roll_alone}
\end{figure}
We verify the GC module on Gaussian clusters whose $k$-NN graph yields a Swiss roll with a sparse section. Algorithm~\ref{alg:graph_connectivity} generates auxiliary points bridging gaps along the manifold, yielding a single connected graph with well-defined geodesic distances (Fig.~\ref{fig:swiss_roll_alone}). Fig.~\ref{fig:line_connection} further validates that diffusion contracts auxiliary points onto the Euclidean geodesic between components, reducing mean perpendicular deviation by $99.2$\% ($\mu_\perp$: $5.16 \to 0.04$) with uniform alignment across the inter-cluster gap.

\subsection{Unbalanced module (UGS)}
Pheno-GS's unbalanced module extends the heat-geodesic Sinkhorn updates to support transport between distributions of differing total mass. Balanced OT requires exact marginal matching ($\boldsymbol{\gamma}\mathbf{1}=\boldsymbol{\mu}$, $\boldsymbol{\gamma}^\top\mathbf{1}=\boldsymbol{\nu}$), rendering it undefined when masses differ. The unbalanced module relaxes this by penalizing marginal deviations via KL divergences~\cite{chizat2018scaling,sejourne2019sinkhorn}:
\begin{align*}
W_p^{\mathrm{UOT}}(\boldsymbol{\mu},\boldsymbol{\nu}) = \biggl(&\inf_{\boldsymbol{\pi} \geq 0} \int_{\mathcal{X}^2} d(x,y)^p \, d\pi(x, y) \\
&+ \tau_{\mu} D(\boldsymbol{\pi}_1 \| \boldsymbol{\mu}) + \tau_{\nu} D(\boldsymbol{\pi}_2 \| \boldsymbol{\nu})\biggr)^{1/p},
\end{align*}
which is finite even when $\boldsymbol{\mu}(\mathcal{X})\neq\boldsymbol{\nu}(\mathcal{X})$~\cite{chizat2018scaling}. As $\tau_\mu,\tau_\nu \to\infty$, UOT recovers classical OT~\cite{pham2020unbalanced,chizat2018scaling}.

Using KL divergence penalties, the entropy-regularized objective~\cite{chizat2018scaling,sejourne2019sinkhorn} is: 
\begin{equation*}
\min_{\boldsymbol{\gamma} \geq 0} \; \varepsilon \, D_{\mathrm{KL}}(\boldsymbol{\gamma} \| \mH_t) + \tau_\mu \, D_{\mathrm{KL}}(\boldsymbol{\gamma} \mathbf{1} \| \boldsymbol{\mu}) + \tau_\nu \, D_{\mathrm{KL}}(\boldsymbol{\gamma}^\top \mathbf{1} \| \boldsymbol{\nu}),
\end{equation*}
where $\mH_t = e^{-tL}$ is the approximated heat kernel~\cite{huguet2023geodesic}, which encodes the heat-geodesic ground cost $C_{ij} = -4t \log (H_t)_{ij}$ implicitly through $\mH_t = e^{-C/4t}$; $\varepsilon$ controls the scale of the reference-divergence term, and $\tau_\mu, \tau_\nu$ control marginal relaxation.
Following~\cite{
pham2020unbalanced}, we use the generalized KL divergence, which does not require normalized inputs and reduces to standard KL when masses match. The GS updates from \eqref{eq:geo_sinkhorn_updates} are modified by exponentiating each diffusion step~\cite{chizat2018scaling,sejourne2019sinkhorn}:
\[
\vv \leftarrow \left[\vmu \oslash p_K(\mL,t)(\va\odot\vw)\right]^{\phi_{v}},
\,
\vw \leftarrow \left[\vnu \oslash p_K(\mL,t)(\va\odot\vv)\right]^{\phi_{w}}
\]
with exponents~\cite{chizat2018scaling}
\[
\phi_v = \frac{\tau_\mu}{\tau_\mu+\varepsilon},
\qquad
\phi_w = \frac{\tau_\nu}{\tau_\nu+\varepsilon} \;\in\;(0,1).
\]

As $\tau_\mu,\tau_\nu \rightarrow \infty$, $\phi \rightarrow 1$ and balanced Sinkhorn are recovered~\cite{cuturi2013sinkhorn}; smaller $\tau$ values progressively relax the marginals. UGS retains GS's per-iteration cost $\mathcal{O}(Kmn)$ for Chebyshev degree K and m-NN graph. The full algorithm is given in Algorithm~\ref{alg:sinkhorn_geo}.
\begin{algorithm}[!t]
    \caption{Unbalanced GS}
    \label{alg:sinkhorn_geo}
\begin{algorithmic}
    \footnotesize
    \STATE {\bfseries Input:} Laplacian $\mL$, distributions $\vmu,\vnu$ on $\gG$, Chebyshev degree $K$, regularization $\lambda = 4t$, marginal penalties $\tau_\mu,\tau_\nu$, vertex weights $\va$
    \STATE {\bfseries Output:} UOT Transport Plan $\boldsymbol{\gamma}^\star$
    \STATE $\vw \leftarrow \mathbf{1}$
    \FOR{$j = 1, 2, \dotsc$}
        \STATE $\vv \leftarrow \left[\vmu \oslash p_K(\mL,t)(\va\odot\vw)\right]^{\phi_{v}}$
        \STATE $\vw \leftarrow \left[\vnu \oslash p_K(\mL,t)(\va\odot\vv)\right]^{\phi_{w}}$
    \ENDFOR
    \STATE {\bfseries Return:} $\boldsymbol{\gamma}^\star = \operatorname{diag}(\boldsymbol{v})\cdot p_K(\boldsymbol{L},t)\cdot \operatorname{diag}(\boldsymbol{w})$
\end{algorithmic}
\end{algorithm}
\begin{figure}[!t]
    \centering
    \includegraphics[width=\linewidth]{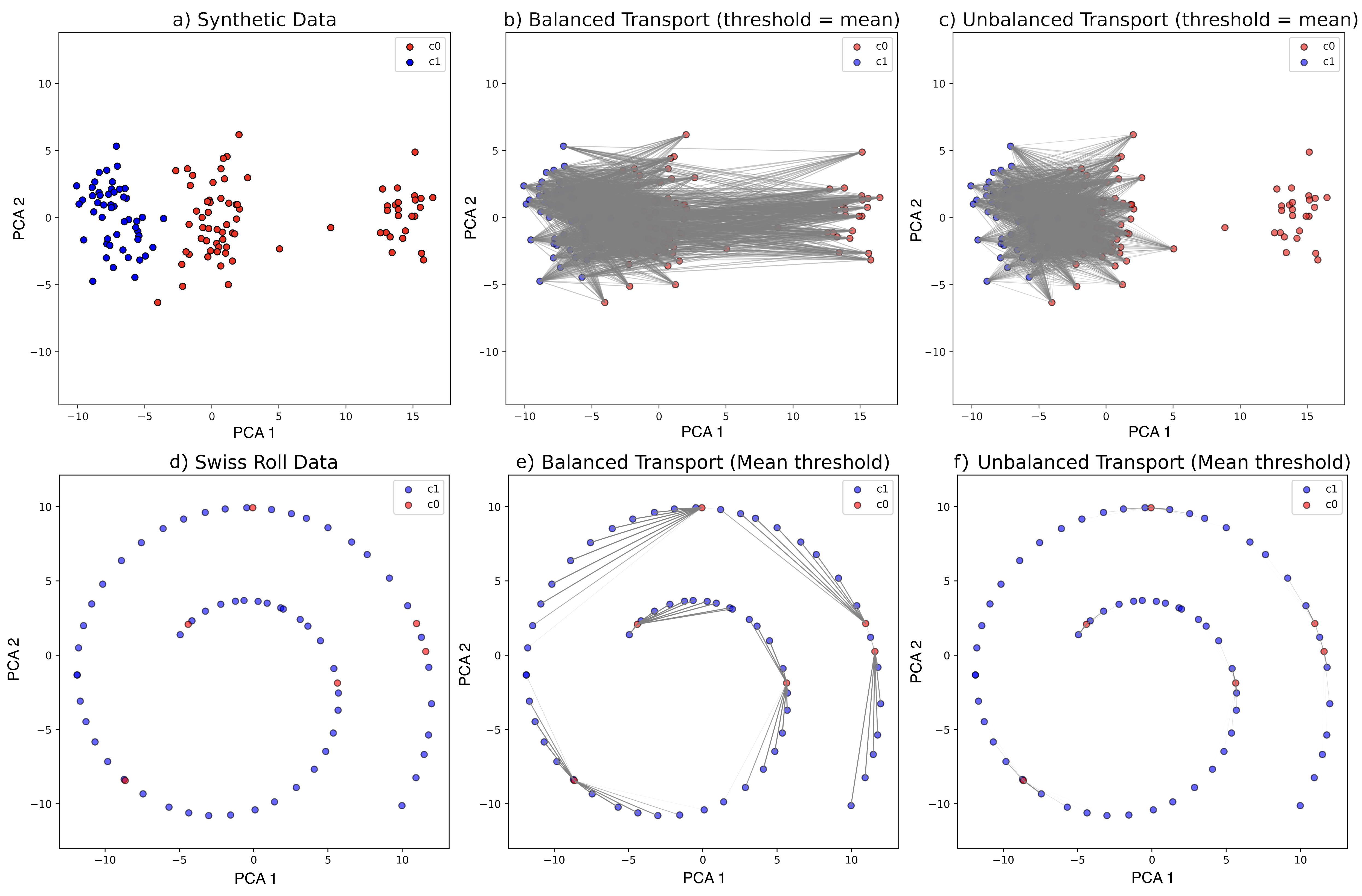}
    \caption{\footnotesize GS vs.\ unbalanced Pheno-GS transport plans. \textbf{(a--c)} Synthetic clusters with outliers: input distributions $c_0$, $c_1$, balanced plan (mass spread across outliers), unbalanced plan (mass concentrated on  main cluster). \textbf{(d--f)} Swiss roll: balanced plan (diffuse, manifold-wide), unbalanced plan (concentrated on geodesically proximal points).}
    \label{fig:transport_between_distributions}
\end{figure}
\begin{figure}[!t]
    \includegraphics[width=\linewidth]{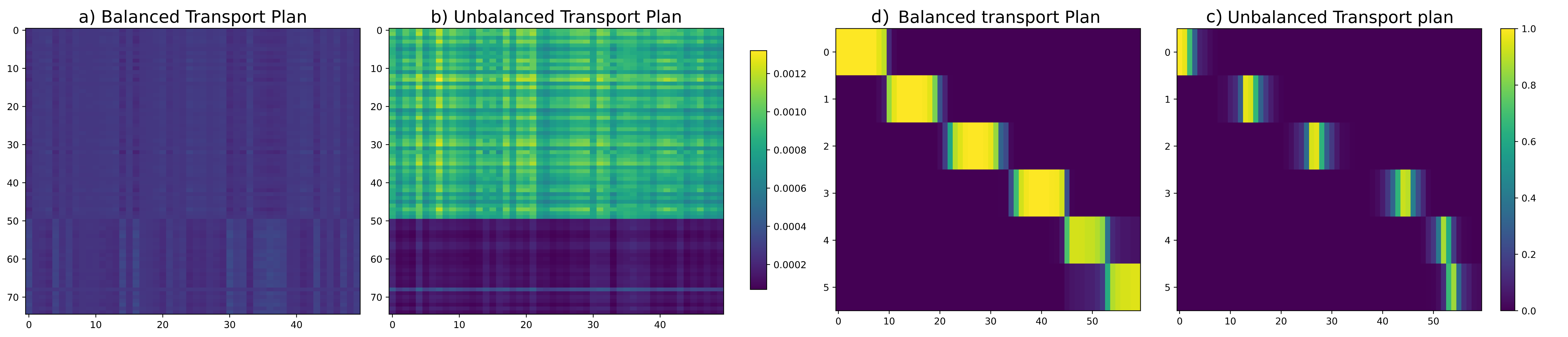}
    \caption{\footnotesize Synthetic cluster transport plans: \textbf{(a)} GS: mass spread across outliers. \textbf{(b)} Unbalanced Pheno-GS: mass concentrated on central cluster. Swiss roll transport plans: \textbf{(c)} GS: broad diagonal reflecting full mass conservation. \textbf{(d)} Unbalanced Pheno-GS: sharp diagonal concentrating mass locally.}
   \label{fig:UOT_OT_hori}
\end{figure}

\subsection{Batched Matrix module (BMS)}
\begin{algorithm}
\caption{Pheno-GS: Batched Matrix module}
\begin{algorithmic}
\STATE \textbf{Input:} Laplacian $\boldsymbol{L}$, partition $\{G_j\}_{j=1}^n$, diffusion time $t$, Chebyshev degree $K$, Chebyshev coefficients $c_0,\dots,c_K$ of the degree-$K$ approximation $p_K(\boldsymbol{L},t)$
\STATE \textbf{Output:} Symmetric pairwise distance matrix $\boldsymbol{D}\in\mathbb{R}^{n\times n}$
\STATE $M_{ij} \leftarrow |G_j|^{-1}\mathbf{1}[i\in G_j], \quad a_0 \leftarrow N^{-1}, \quad \boldsymbol{a} \leftarrow a_0 \mathbf{1}_N$
\STATE \textit{//Rescaling\footnotemark}
\STATE $\boldsymbol{\xi} \leftarrow \boldsymbol{a} \left(\tfrac{1}{2}c_0 + \sum_{i=1}^{K} (-1)^i c_i \right)$
\FOR{$j = 1,\dots,n$}
    \STATE $u_j \leftarrow \sum_{i\in G_j} M_{ij}\ln(M_{ij}/\xi_i)$
    \STATE $R_{ij} \leftarrow a_0 M_{ij}/\xi_i \ \ (i\in G_j), \quad R_{ij} \leftarrow 0 \ \ (i\notin G_j)$
\ENDFOR
\STATE \textit{// Second diffusion}
\STATE $\boldsymbol{Q} \leftarrow p_K(\boldsymbol{L},t)\,\boldsymbol{R}$
\FOR{$1 \le j < k \le n$}
    \STATE $w_{jk} \leftarrow \sum_{i\in G_k} M_{ik}\ln(M_{ik}/Q_{ij})$
\ENDFOR
\STATE $D_{jk} \leftarrow 4a_0 t\,(w_{jk} + u_j) \quad \forall j<k$
\STATE $\boldsymbol{D} \leftarrow \boldsymbol{D} + \boldsymbol{D}^\top$
\STATE \textbf{Return:} $\boldsymbol{D}$
\end{algorithmic}
\end{algorithm}
\footnotetext{This rescaling assumes $\boldsymbol{L}\mathbf{1} = 0$ (combinatorial Laplacian); for the normalized Laplacian, use $\boldsymbol{\xi} \leftarrow p_K(\boldsymbol{L},t)\,\boldsymbol{a}$ instead.}

Pheno-GS's batched matrix module computes the full $n\times n$ pairwise distance matrix in a single vectorized operation. Rather than $\binom{n}{2}$ sequential Sinkhorn runs each requiring $2T_{\text{iter}}$ heat kernel applications, it performs one heat diffusion over all distributions simultaneously, regardless of $n$. The full distance matrix follows from the matrix-level diffusion at cost $\mathcal{O}(N \cdot n)$. 

Given $n$ distributions partitioning $N$ nodes of a common graph into disjoint supports $\{G_j\}_{j=1}^n$, we define the $N\times n$ distribution matrix
\[
M_{ij} = \begin{cases} |G_j|^{-1}, & i \in G_j,\\ 
0, & \text{otherwise,}\end{cases}
\]
so each column $M_{\cdot,j}$ is a distribution on $G_j$, with $a_0 = N^{-1}$ and $\va = a_0\mathbf{1}_N$. We rescale the uniform background measure by its closed-form diffusion coefficient to obtain a geometry-only baseline $\boldsymbol{\xi}$, and compute the self-divergence $\boldsymbol{u}$ and ratio field for each distribution $\boldsymbol{R}$. $Rij$ is the source-side scaling obtained from the first heat diffusion $\boldsymbol{\xi}$. $\boldsymbol{u}$ is the self-divergence term:
\[
u_j \leftarrow \sum_{i \in G_j} M_{ij} \ln \frac{M_{ij}}{\xi_i},  \qquad R_{ij} = a_0\,M_{ij}/\xi_i.
\]
Applying the heat kernel to all columns of $R$ simultaneously, $\mQ \leftarrow p_K(\mL,\,t)\,\mR,$ propagates every distribution across the manifold in a single batched operation. $Q_{ij}$ encodes how much signal from distribution $j$ reaches node $i$. 
The batched module therefore reproduces the pairwise distances GS computes, to the numerical precision of the shared Chebyshev kernel, while replacing the ($n$ choose $2$) separate diffusions with a single pass. The cross-divergence $\boldsymbol{w}$ and the full $n\times n$ matrix is then assembled as
\[
w_{jk} \leftarrow \sum_{i \in G_k} M_{ik}\ln\frac{M_{ik}}{Q_{ij}},
\qquad
D_{jk} \leftarrow 4a_0 t\,(w_{jk} + u_j), \quad j < k,
\]
symmetrized as $\mD \leftarrow \mD + \mD^\top$. Pheno-GS produces results identical to GS up to numerical precision from the shared Chebyshev approximation $p_K(\mL, t)$.

\begin{table*}[!t]
    \centering
    \caption{\footnotesize KNN task results for $15$ distributions (mean $\pm$ std over runs). Best scores in \textbf{bold}.}
    \label{tab:knn_results_new}
{\footnotesize
\begin{tabular}{lcccccc}
\toprule
Method & SpearmanR & PearsonR & $P@5$ & $P@10$ & Time (s) w/o graph & Time (s) \\
\midrule
Diffusion EMD      & 0.628 $\pm$ 0.099\textsuperscript{d} & 0.739 $\pm$ 0.029\textsuperscript{e} & 0.657 $\pm$ 0.074\textsuperscript{b} & 0.772 $\pm$ 0.035\textsuperscript{b} & 0.920 $\pm$ 0.038\textsuperscript{b} & 2.237 $\pm$ 0.148\textsuperscript{b} \\
Euler Sinkhorn     & 0.701 $\pm$ 0.076\textsuperscript{c} & 0.768 $\pm$ 0.061\textsuperscript{abcde} & 0.665 $\pm$ 0.064\textsuperscript{b} & 0.827 $\pm$ 0.037\textsuperscript{a} & 35.374 $\pm$ 1.539\textsuperscript{g} & 36.926 $\pm$ 1.464\textsuperscript{i} \\
Sinkhorn $W_1$     & 0.385 $\pm$ 0.046\textsuperscript{f} & 0.482 $\pm$ 0.042\textsuperscript{g} & 0.467 $\pm$ 0.028\textsuperscript{d} & 0.725 $\pm$ 0.027\textsuperscript{c} & 16.955 $\pm$ 0.310\textsuperscript{e} & 16.955 $\pm$ 0.310\textsuperscript{g} \\
Sinkhorn $W_2$     & 0.385 $\pm$ 0.046\textsuperscript{f} & 0.482 $\pm$ 0.042\textsuperscript{g} & 0.467 $\pm$ 0.028\textsuperscript{d} & 0.725 $\pm$ 0.027\textsuperscript{c} & 17.249 $\pm$ 0.987\textsuperscript{e} & 17.249 $\pm$ 0.987\textsuperscript{g} \\
LR Sinkhorn $W_2$  & -0.384 $\pm$ 0.094\textsuperscript{g} & -0.311 $\pm$ 0.082\textsuperscript{h} & 0.239 $\pm$ 0.056\textsuperscript{e} & 0.648 $\pm$ 0.017\textsuperscript{d} & 182.162 $\pm$ 33.942\textsuperscript{h} & 182.162 $\pm$ 33.942\textsuperscript{j} \\
UOT Sinkhorn       & 0.396 $\pm$ 0.042\textsuperscript{e} & 0.525 $\pm$ 0.036\textsuperscript{f} & 0.481 $\pm$ 0.030\textsuperscript{c} & 0.725 $\pm$ 0.025\textsuperscript{c} & 19.146 $\pm$ 0.899\textsuperscript{f} & 19.146 $\pm$ 0.899\textsuperscript{h} \\
GS                 & 0.849 $\pm$ 0.043\textsuperscript{ab} & {0.790 $\pm$ 0.015}\textsuperscript{b} & \textbf{0.807 $\pm$ 0.047}\textsuperscript{a} & {0.848 $\pm$ 0.041}\textsuperscript{a} & 1.536 $\pm$ 0.103\textsuperscript{c} & 2.886 $\pm$ 0.146\textsuperscript{c} \\
\midrule
Pheno-GS (BMS)         & 0.849 $\pm$ 0.043\textsuperscript{ab} & {0.790 $\pm$ 0.015}\textsuperscript{b} & \textbf{0.807 $\pm$ 0.047}\textsuperscript{a} & {0.848 $\pm$ 0.041}\textsuperscript{a} & 0.473 $\pm$ 0.101\textsuperscript{a} & \textbf{1.907 $\pm$ 0.212}\textsuperscript{a} \\
Pheno-GS (GC + BMS)    & 0.845 $\pm$ 0.038\textsuperscript{b} & 0.788 $\pm$ 0.016\textsuperscript{c} & 0.797 $\pm$ 0.049\textsuperscript{a} & 0.845 $\pm$ 0.031\textsuperscript{a} & \textbf{0.420 $\pm$ 0.071}\textsuperscript{a} & 5.900 $\pm$ 0.393\textsuperscript{e} \\
Pheno-GS (UGS)         & 0.840 $\pm$ 0.042\textsuperscript{b} & \textbf{0.793 $\pm$ 0.015}\textsuperscript{a} & {0.799 $\pm$ 0.038}\textsuperscript{a} & \textbf{0.853 $\pm$ 0.040}\textsuperscript{a} & 1.655 $\pm$ 0.169\textsuperscript{d} & 3.053 $\pm$ 0.218\textsuperscript{d} \\
Pheno-GS (GC + UGS)    & \textbf{0.850 $\pm$ 0.037}\textsuperscript{a} & 0.786 $\pm$ 0.015\textsuperscript{d} & 0.805 $\pm$ 0.043\textsuperscript{a} & 0.846 $\pm$ 0.030\textsuperscript{a} & 1.691 $\pm$ 0.067\textsuperscript{d} & 7.009 $\pm$ 0.216\textsuperscript{f} \\
\bottomrule
\end{tabular}%
}
\vspace{2pt}
\footnotesize{Superscript letters denote groupings from paired Wilcoxon signed-rank tests ($\alpha=0.05$, uncorrected) within each column; methods sharing a letter are not significantly different from one another. GS-vs-Pheno-GS comparisons reported in the text use a separate Holm correction across the four Pheno-GS variants per metric.}
\end{table*}

\noindent \section{Empirical Results}\vspace {-2mm}
We evaluate Pheno-GS robustness under sparse and disconnected manifold structure with mass-imbalanced distributions and scalability to large collections of distributions.

\begin{figure}[!t]
    \centering
    \includegraphics[width=\linewidth]{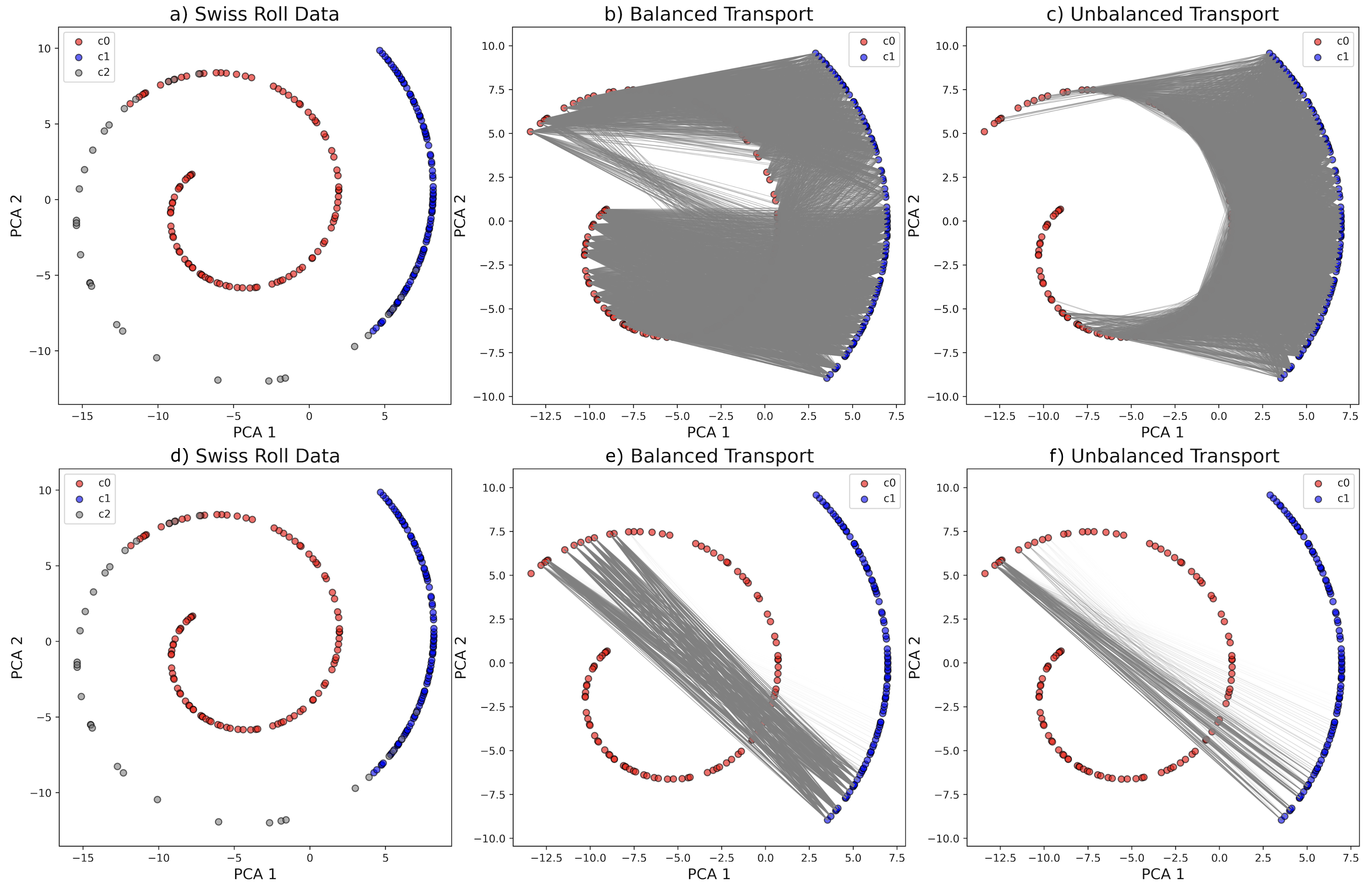}
    \caption{\footnotesize Effect of graph connectivity on transport. \textbf{(a--c)} Without connectivity: three distributions ($c_0$, $c_1$, sparse bridge $c_2$), balanced plan (ignores manifold), unbalanced plan (distorted geodesics). \textbf{(d--f)} With connectivity: same distributions, balanced plan (geodesically consistent), unbalanced plan (geodesic alignment with suppression of distant pairs).}
    \label{fig:non_onnected_swiss_roll_final}
\end{figure}
\begin{figure}[!t]
    \centering
    \includegraphics[width=\linewidth]{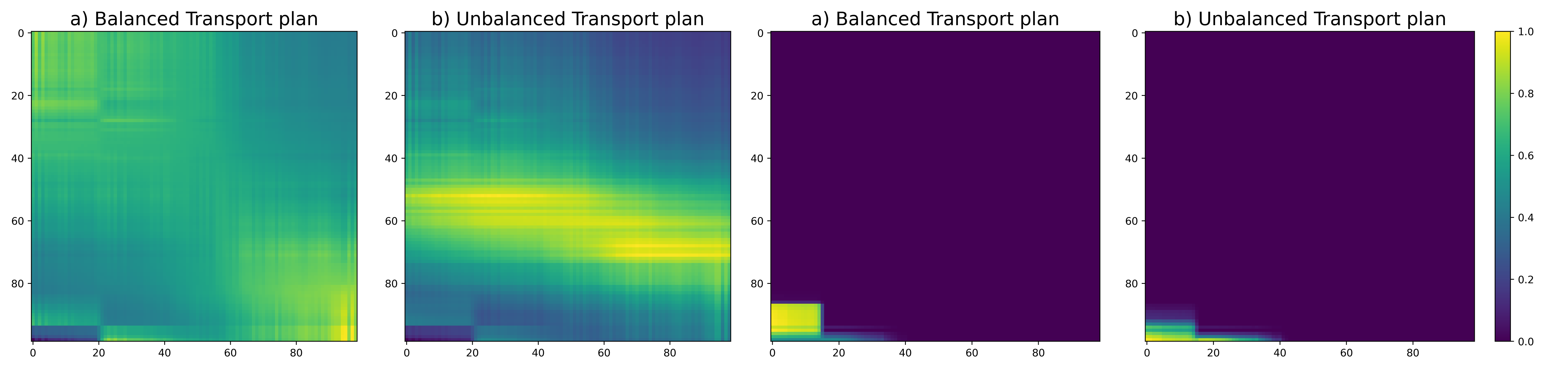}
    \caption{\footnotesize Without graph connectivity: \textbf{(a)} Balanced: mass spread across all pairs, ignoring manifold structure. \textbf{(b)} Unbalanced: distorted by bandwidth-induced Euclidean-like geodesics. With graph connectivity: \textbf{(c)} Balanced: mass peaks at geodesically adjacent pairs. \textbf{(d)} Unbalanced: sharper local concentration with greater suppression of distant pairs.}
    \label{fig:warped_swiss_roll_transport_transport_final}
\end{figure}
\textbf{Graph Connectivity and Robustness} We first validate qualitatively on a Swiss roll with three regions: $c_0$ ($100$ points, inner arm), $c_1$ ($100$ points, outer arm), and a sparse bridge $c_2$ (25 points). Without 
the graph connectivity module the $k$-NN graph is disconnected. Naively widening the kernel bandwidth connects components but introduces spurious shortcuts that distort manifold geometry. With GC enabled and applied to the sparse bridge, both modules recover geometrically consistent plans. The batched module routes mass between $c_0$ and $c_1$ via the sparse bridge, while the unbalanced module further concentrates transport on geodesically proximal pairs (Figures~\ref{fig:non_onnected_swiss_roll_final}--\ref{fig:warped_swiss_roll_transport_transport_final}). Figure~\ref{fig:transport_between_distributions} compares GS with Pheno-GS's unbalanced module on synthetic data with outliers and on a Swiss roll. GS transports all mass including outliers, distributing it broadly across the manifold and producing dense plans. Pheno-GS concentrates transport geodesically, suppressing distant mass and yielding more localized plans~\cite{chizat2018scaling,sejourne2019sinkhorn} (Figure~\ref{fig:UOT_OT_hori}).

We then benchmark quantitatively against Euclidean Sinkhorn~\cite{cuturi2013sinkhorn}, LR Sinkhorn~\cite{scetbon2021low}, DiffusionEMD~\cite{tong2021diffusion}, and Euler Sinkhorn on a $k$-nearest-neighbor task. Following~\cite{huguet2023geodesic}, we sample $15$ Gaussian distributions on a Swiss roll ($2000$ observations per distribution, embedded in $10$ dimensions via random rotation) and evaluate against ground-truth geodesic distances, reporting Spearman/Pearson correlations, $P@5$ and $P@10$ averaged over 10 seeds (Table~\ref{tab:knn_results_new}). Pheno-GS (BMS) reproduces GS's distances exactly while running substantially faster; the unbalanced and connectivity variants match GS to within noise. All Pheno-GS configurations substantially outperform the Euclidean- and low-rank-Sinkhorn baselines, and do so orders of magnitude faster than Euler and LR Sinkhorn.
\begin{table*}[!t]
    \centering
    \caption{\footnotesize KNN task results (mean $\pm$ std over runs). Best scores in \textbf{bold}.}
    \label{tab:knn_results_new_MGS}
{\footnotesize
\begin{tabular}{lcccccc}
\toprule
Method & SpearmanR & PearsonR & $P@5$ & $P@10$ & Time (s) w/o graph & Time (s) \\
\midrule
Diffusion EMD      & 0.571 $\pm$ 0.042\textsuperscript{c} & 0.650 $\pm$ 0.016\textsuperscript{b} & 0.557 $\pm$ 0.044\textsuperscript{b} & 0.582 $\pm$ 0.021\textsuperscript{b} & 5.234 $\pm$ 0.065\textsuperscript{b} & 6.991 $\pm$ 0.095\textsuperscript{b} \\
Euler Sinkhorn     & 0.642 $\pm$ 0.057\textsuperscript{b} & 0.673 $\pm$ 0.054\textsuperscript{b} & 0.467 $\pm$ 0.041\textsuperscript{d} & 0.530 $\pm$ 0.024\textsuperscript{c} & 553.645 $\pm$ 16.397\textsuperscript{e} & 555.394 $\pm$ 16.376\textsuperscript{e} \\
Sinkhorn $W_1$     & 0.327 $\pm$ 0.028\textsuperscript{d} & 0.388 $\pm$ 0.025\textsuperscript{c} & 0.480 $\pm$ 0.042\textsuperscript{c} & 0.478 $\pm$ 0.027\textsuperscript{d} & 27.147 $\pm$ 0.238\textsuperscript{d} & 27.147 $\pm$ 0.238\textsuperscript{d} \\
Sinkhorn $W_2$     & 0.327 $\pm$ 0.028\textsuperscript{d} & 0.388 $\pm$ 0.025\textsuperscript{c} & 0.480 $\pm$ 0.042\textsuperscript{c} & 0.478 $\pm$ 0.027\textsuperscript{d} & 27.147 $\pm$ 0.126\textsuperscript{d} & 27.147 $\pm$ 0.126\textsuperscript{d} \\
LR Sinkhorn $W_2$  & -0.293 $\pm$ 0.050\textsuperscript{e} & -0.292 $\pm$ 0.039\textsuperscript{d} & 0.056 $\pm$ 0.021\textsuperscript{e} & 0.144 $\pm$ 0.026\textsuperscript{e} & 750.451 $\pm$ 1.561\textsuperscript{f} & 750.451 $\pm$ 1.561\textsuperscript{f} \\
GS                 & \textbf{0.824 $\pm$ 0.014}\textsuperscript{a} & \textbf{0.723 $\pm$ 0.009}\textsuperscript{a} & \textbf{0.784 $\pm$ 0.039}\textsuperscript{a} & \textbf{0.810 $\pm$ 0.016}\textsuperscript{a} & 18.152 $\pm$ 0.286\textsuperscript{c} & 19.918 $\pm$ 0.271\textsuperscript{c} \\
\midrule
Pheno-GS (BMS)     & \textbf{0.824 $\pm$ 0.014}\textsuperscript{a} & \textbf{0.723 $\pm$ 0.009}\textsuperscript{a} & \textbf{0.784 $\pm$ 0.039}\textsuperscript{a} & \textbf{0.810 $\pm$ 0.016}\textsuperscript{a} & \textbf{0.570 $\pm$ 0.221}\textsuperscript{a} & \textbf{2.319 $\pm$ 0.210}\textsuperscript{a} \\
\bottomrule
\end{tabular}%
}
\vspace{2pt}
\footnotesize{Superscript letters denote groupings from paired Wilcoxon signed-rank tests ($\alpha=0.05$, uncorrected) within each column; methods sharing a letter are not significantly different from one another. GS-vs-Pheno-GS comparisons reported in the text use a separate Holm correction.}
\end{table*}
\begin{table}[!t]
\centering
\caption{\footnotesize Runtime (s) of Pheno-GS (BMS) for computing the pairwise distance matrix with increasing number of distributions $n$ each with $5000$ observations. Speedup is relative to GS.}
\label{tab:runtime_comparison}
{\footnotesize
\begin{tabular}{lccccc}
\toprule
$n$ & $100$ & $200$ & $300$ & $400$ & $500$ \\
\midrule
GS & 196.71 & 1891.10 & 6093.75 & 15330.76 & 29341.81 \\
Pheno-GS  & \textbf{7.97} & \textbf{27.45} & \textbf{53.02} & \textbf{89.05} & \textbf{137.61} \\
\midrule
Speedup           & $24.7\times$ & $68.9\times$ & $114.9\times$ & $172\times$ & $213\times$ \\
\bottomrule
\end{tabular}%
}
\end{table}

\textbf{Scalability Assessment} We verify that the batched matrix module produces distances numerically consistent with GS. On a $250$ distribution synthetic dataset~\cite{zapatero2022pdos}, the distance matrices showed near-perfect agreement: MAE ($=3.60\times10^{-12}$), RMSE ($=4.58\times10^{-12}$), Pearson ($r=1.000$), with negligible bias ($-3.47\times10^{-12}$), confirming that the batched matrix module introduces no meaningful numerical error beyond the Chebyshev kernel approximation shared by both methods. This agreement is measured with GS and BMS operating on the identical graph and Chebyshev parameters.

We then benchmark scalability on synthetic data ($5000$ points and $35$ features per distribution) as the number of distributions increases from $100$ to $500$. With a fixed number of cells \(c\) per distribution (so total nodes \(N = cn\)), the batched module amortizes a single shared diffusion across all pairs rather than running \(\binom{n}{2}\) separate solves. Empirically, its runtime scales as \(\approx n^{1.8}\) versus \(\approx n^{3.1}\) for sequential GS (Table~\ref{tab:runtime_comparison}), a gap that widens with \(n\), reaching a \(213\times\) speedup at \(n=500\).

 We evaluate accuracy against GS, Sinkhorn~\cite{cuturi2013sinkhorn}, LR Sinkhorn~\cite{scetbon2021low}, DiffusionEMD~\cite{tong2021diffusion}, and Euler Sinkhorn~\cite{solomon2015convolutional} on a $k$-nearest-neighbor task. Following~\cite{huguet2023geodesic}, we sample $50$ Gaussian distributions on a Swiss roll ($1000$ observations per distribution, embedded in $10$ dimensions) and evaluate via Spearman/Pearson correlation, $P@5$ and $P@10$ (Table~\ref{tab:knn_results_new_MGS}). The batched matrix module reproduces GS's accuracy exactly while running $16 \times$ faster excluding graph construction, and outperforms all Euclidean-based baselines by a substantial margin.

\begin{figure}[!t]
    \centering
    \includegraphics[width=\linewidth]{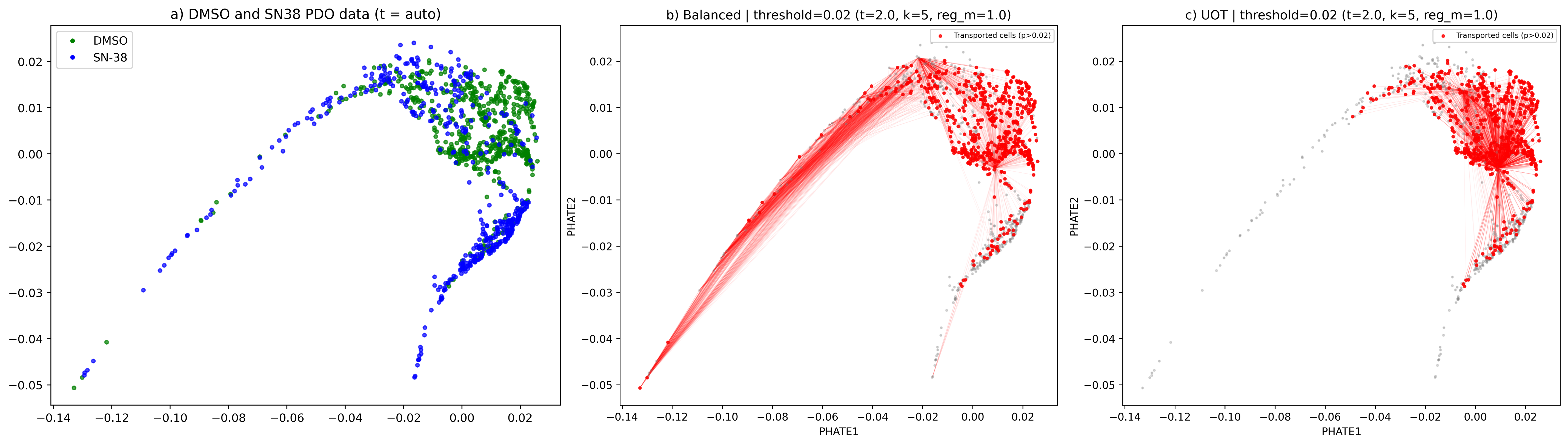}
    \caption{\footnotesize GS vs.\ Unbalanced Pheno-GS from DMSO to SN-38 on a PDO CyTOF dataset. a) PHATE embedding colored by condition. b) Balanced transport: transport mass $ > 0.02$ spans the full manifold, forced by strict marginal constraints. c) Unbalanced transport: edges concentrate on phenotypically proximal cell pairs. Cells close to the SN-38 manifold are preferentially transported, reflecting survival likelihood under perturbation; phenotypically distant cells receive near-zero mass, consistent with apoptotic fate.}
    \label{fig:uot_trellis}
\end{figure}

\begin{figure}[!t]
    \centering
    \includegraphics[width=\linewidth]{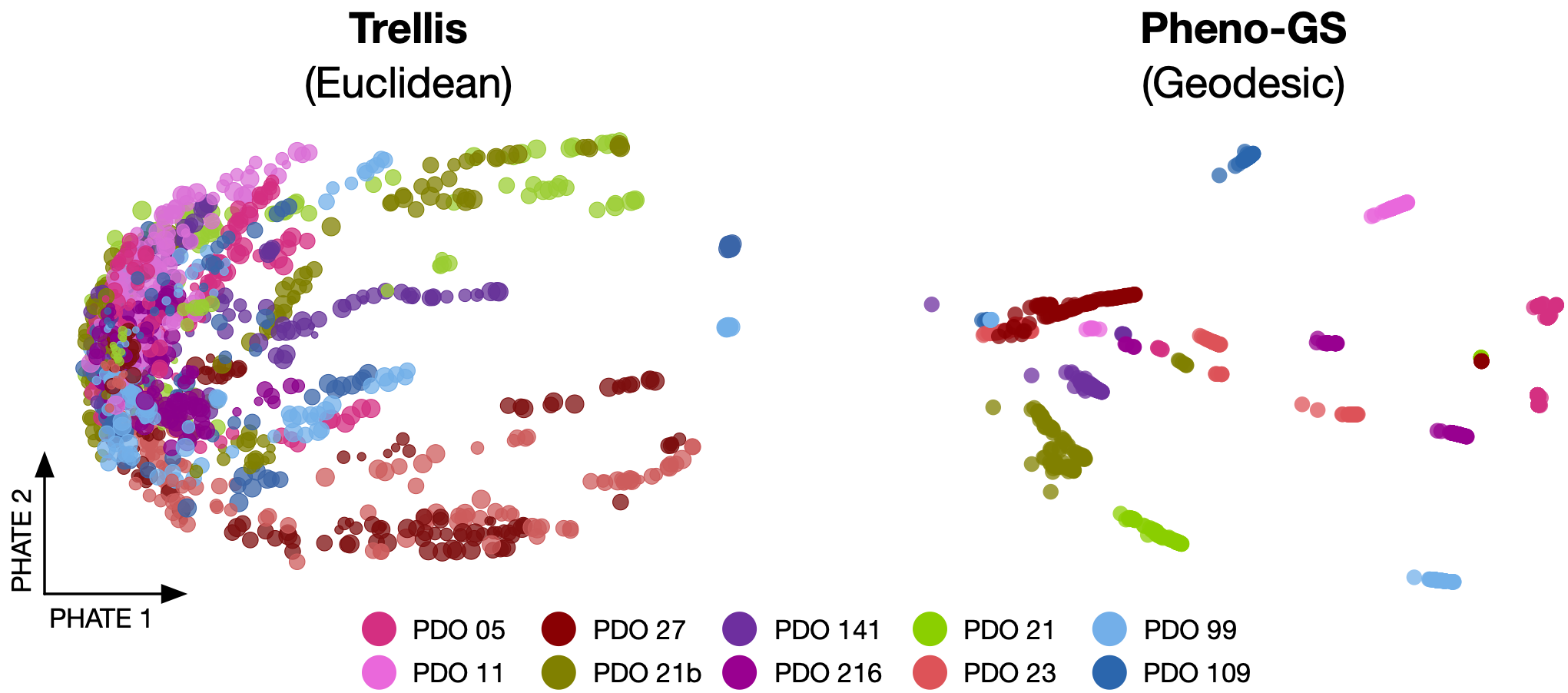}
    \caption{\footnotesize  PHATE Phenoscape embeddings of pairwise OT distances between $>1,500$ PDO treatment distributions ($1$ dot = $1$ perturbation) \cite{zapatero2022pdos} (Left) Trellis (Euclidean Sinkhorn) computes distances relative to a shared control, collapsing all conditions onto a common reference frame and obscuring inter-distribution geometry. (Right) Pheno-GS computes the full pairwise geodesic distance matrix, resolving patient-specific treatment responses as geometrically separated clusters.}
    \label{fig:trellis_v_pheno_GS}
\end{figure}

\textbf{Biological Data} 
We compare Unbalanced Pheno-GS with GS on a PDO CyTOF dataset ~\cite{zapatero2022pdos}. We calculate the transport plan between a DMSO control and SN-38 perturbed condition. GS distributes mass across the full manifold, including long-range connections to phenotypically distant cells, due to strict marginal constraints. Pheno-GS instead concentrates high-probability edges on geodesically proximal cell pairs: DMSO cells whose marker expression most closely resembles the SN-38-perturbed state. This selectivity has a natural biological interpretation: cells already phenotypically similar to the drug-adapted state are most likely to survive perturbation and undergo transition, while phenotypically distant cells are effectively excluded from transport. Rather than forcing these cells into spurious long-range assignments, the KL marginal penalties allow their mass to be absorbed, providing a principled mechanism to model apoptosis within the transport framework. Pheno-GS thus yields biologically coherent plans that distinguish phenotypic transition from cell death.

Finally, we apply Pheno-GS to a biological dataset containing $>1,500$ distributions of cancer cells (Figure \ref{fig:trellis_v_pheno_GS}). Each dataset in this population is a patient-derived cancer-organoid (PDO) dataset from \cite{zapatero2022pdos} that has been treated with a collection of drugs and co-culturing conditions. Here we compare Trellis, a balanced tree-based OT, with Pheno-GS. Pheno-GS recovers patient identity as the dominant source of distributional variation, with treatment-induced perturbations visible as structured displacement within each patient cluster. Across the 10 PDOs ($\text{chance} \approx 0.10$), both methods recovered identity locally (kNN $k=1$: old $0.754$, new $0.996$; $p \approx 0.001$). Globally, Pheno-GS also yields better cluster separation than Trellis, flipping the silhouette from $-0.099$ to $+0.106$.
\section{Conclusion}\vspace{-2mm}
We presented Pheno-GS, a manifold OT system that addresses three fundamental limitations of GS:  $1$) the graph connectivity module guarantees well-defined geodesic distances on sparse or disconnected manifolds; $2$) the unbalanced module supports transport between distributions of differing total mass; and $3$) the batched matrix module reduces pairwise computation from $\mathcal{O}\!\left(\binom{n}{2}\cdot T_{\mathrm{iter}}\right)$ to $\mathcal{O}(N \cdot n)$, achieving over $200\times$ speedup for $n = 500$ with no loss in accuracy relative to GS. Pheno-GS therefore provides a principled and scalable framework for OT on complex, real-world phenoscape-scale manifolds.



\let\OLDthebibliography\thebibliography
\renewcommand\thebibliography[1]{
  \OLDthebibliography{#1}
  \scriptsize
  \setlength{\itemsep}{0pt}
  \setlength{\parskip}{0pt}
}
\bibliographystyle{IEEEbib}
\bibliography{tidy}
\end{document}

%% file: math_commads.tex
\usepackage{amsmath,amsfonts,bm}
\usepackage{soul} 

\def\eqref#1{equation~\ref{#1}}

\def\1{\bm{1}}

\def\vmu{{\bm{\mu}}}
\def\vnu{{\bm{\nu}}} 

\def\va{{\bm{a}}}

\def\vv{{\bm{v}}}
\def\vw{{\bm{w}}}

\def\mA{{\bm{A}}}

\def\mD{{\bm{D}}}

\def\mH{{\bm{H}}}
\def\mI{{\bm{I}}}

\def\mK{{\bm{K}}}
\def\mL{{\bm{L}}}

\def\mN{{\bm{N}}}

\def\mP{{\bm{P}}}
\def\mQ{{\bm{Q}}}
\def\mR{{\bm{R}}}
\def\mS{{\bm{S}}}

\def\mX{{\bm{X}}}

\DeclareMathAlphabet{\mathsfit}{\encodingdefault}{\sfdefault}{m}{sl}
\SetMathAlphabet{\mathsfit}{bold}{\encodingdefault}{\sfdefault}{bx}{n}

\def\gG{{\mathcal{G}}}



%% file: tidy.bib
@article{chen2020uncovering,
	title        = {Uncovering axes of variation among single-cell cancer specimens},
	author       = {Chen, William S and Zivanovic, Nevena and Van Dijk, David and Wolf, Guy and Bodenmiller, Bernd and Krishnaswamy, Smita},
	year         = 2020,
	journal      = {Nat. methods},
	publisher    = {Nature Publishing Group}
}

@article{crane_geodesics_2013,
	title        = {Geodesics in heat},
	author       = {Crane, K. and Weischedel, C. and Wardetzky, M.},
	year         = 2013,
	journal      = {ACM Trans. Graphics},
	volume       = 32,
	number       = 5,
	pages        = {1--11}
}

@article{cuturi2013sinkhorn,
	title        = {Sinkhorn distances: Lightspeed computation of optimal transport},
	author       = {Cuturi, M.},
	year         = 2013,
	journal      = {NeurIPS},
	volume       = 26
}

@inproceedings{marcotte2022fast,
	title        = {Fast multiscale diffusion on graphs},
	author       = {Marcotte, S. and Barb{\'e}, A. and Gribonval, R. and others},
	year         = 2022,
	booktitle    = {Proc. ICASSP},
	organization = {IEEE},
	pages        = {5627--5631}
}

@inproceedings{scetbon2021low,
	title        = {Low-rank {Sinkhorn} factorization},
	author       = {Scetbon, M. and Cuturi, M. and Peyr{\'e}, G.},
	year         = 2021,
	booktitle    = {Proc. ICML},
	pages        = {9344--9354},
	organization = {PMLR}
}

@inproceedings{shuman_chebyshev_2011,
	title        = {Chebyshev polynomial approximation for distributed signal processing},
	author       = {Shuman, D. I. and Vandergheynst, P. and Frossard, P.},
	year         = 2011,
	booktitle    = {Proc. DCOSS},
	pages        = {1--8},
	organization = {IEEE}
}

@article{solomon2015convolutional,
	title        = {Convolutional {Wasserstein} distances},
	author       = {Solomon, J. and De Goes, F. and Peyr{\'e}, G. and others},
	year         = 2015,
	journal      = {ACM Trans. Graphics},
	volume       = 34,
	number       = 4,
	pages        = {1--11}
}

@inproceedings{tong2021diffusion,
	title        = {Diffusion {Earth} mover's distance and distribution embeddings},
	author       = {Tong, A. and Huguet, G. and Natik, A. and MacDonald, K. and Kuchroo, M. and Coifman, R. and Wolf, G. and Krishnaswamy, S.},
	year         = 2021,
	booktitle    = {Proc. ICML},
	pages        = {10336--10346},
	organization = {PMLR}
}

@article{varadhan_behavior_1967,
	title        = {On the behavior of the fundamental solution of the heat equation with variable coefficients},
	author       = {Varadhan, S. R. S.},
	year         = 1967,
	journal      = {Commun. Pure Appl. Math.},
	volume       = 20,
	number       = 2,
	pages        = {431--455}
}

@article{zapatero2022pdos,
	title        = {Trellis tree-based analysis reveals stromal regulation of patient-derived organoid drug responses},
	author       = {Ramos Zapatero, M. and Tong, A. and Opzoomer, J. W. and others},
	year         = 2023,
	journal      = {Cell},
	volume       = 186,
	number       = 25,
	pages        = {5606--5619}
}

@inproceedings{pham2020unbalanced,
	title        = {On unbalanced optimal transport: An analysis of {Sinkhorn} algorithm},
	author       = {Pham, K. and Le, K. and Ho, N. and Pham, T. and Bui, H.},
	booktitle    = {Proc. ICML},
	pages        = {7673--7682},
	year         = {2020},
	organization = {PMLR}
}

@inproceedings{huguet2023geodesic,
	title        = {Geodesic {Sinkhorn} for fast and accurate optimal transport on manifolds},
	author       = {Huguet, G. and Tong, A. and Ramos Zapatero, M. and Tape, C. J. and Wolf, G. and Krishnaswamy, S.},
	booktitle    = {Proc. MLSP},
	pages        = {1--6},
	year         = {2023},
	organization = {IEEE}
}

@article{chizat2018scaling,
	title        = {Scaling algorithms for unbalanced optimal transport problems},
	author       = {Chizat, L. and Peyr{\'e}, G. and Schmitzer, B. and Vialard, F.-X.},
	journal      = {Math. Comput.},
	volume       = {87},
	number       = {314},
	pages        = {2563--2609},
	year         = {2018}
}

@article{sejourne2019sinkhorn,
	title        = {Sinkhorn divergences for unbalanced optimal transport},
	author       = {S{\'e}journ{\'e}, T. and Feydy, J. and Vialard, F.-X. and Trouv{\'e}, A. and Peyr{\'e}, G.},
	journal      = {arXiv:1910.12958},
	year         = {2019}
}

@inproceedings{dvurechensky_computational_2018,
  title = {Computational {{Optimal Transport}}: {{Complexity}} by {{Accelerated Gradient Descent Is Better Than}} by {{Sinkhorn}}'s {{Algorithm}}},
  shorttitle = {Computational {{Optimal Transport}}},
  author = {Dvurechensky, Pavel and Gasnikov, Alexander and Kroshnin, Alexey},
  year = {2018},
  booktitle = {Proceedings of the 35th {{International Conference}} on {{Machine Learning}}},
  volume = {80},
  pages = {1367--1376},
  publisher = {PMLR},
  series = {Proceedings of {{Machine Learning Research}}},
  editor = {Dy, Jennifer and Krause, Andreas},
  address = {Stockholm, Sweden},
  langid = {english},
}
